\documentclass[11pt]{article}

\PassOptionsToPackage{hyperfootnotes=false}{hyperref}
\usepackage[final]{acl}
\usepackage{times}
\usepackage{latexsym}
\usepackage[T1]{fontenc}
\usepackage[utf8]{inputenc}
\usepackage{microtype}
\usepackage{inconsolata}
\usepackage{booktabs}
\usepackage{array}
\usepackage{amsmath}
\usepackage{amssymb}
\usepackage{graphicx}

\hypersetup{pdftitle={Remember, Verify, or Ask? Evaluating Memory Decisions at Reuse Time},pdfauthor={Baichuan Li; Junyi Yao; Zihao Zheng}}

\newcommand{\mcb}{\textsc{MCB}}
\newcommand{\act}{\textsc{MCB-Act}}

\title{Remember, Verify, or Ask? Evaluating Memory Decisions at Reuse Time}

\author{
Baichuan Li\textsuperscript{1,}\thanks{Baichuan Li, Junyi Yao, and Zihao Zheng contributed equally and share first authorship.}\quad
Junyi Yao\textsuperscript{2,*}\quad
Zihao Zheng\textsuperscript{2,*} \\
\textsuperscript{1}Southern Methodist University \\
Dallas, Texas, USA \\
\textsuperscript{2}Washington University in St. Louis \\
St. Louis, Missouri, USA \\
\texttt{baichuanl@smu.edu} \\
\texttt{j.yao@wustl.edu; z.zihaogary@wustl.edu}
}

\begin{document}
\maketitle

\begin{abstract}
Persistent memory can personalize an LLM agent, but an incorrect durable update can distort future behavior. We introduce the Memory--Clarification Boundary benchmark (MCB), which supplies an interaction-derived candidate and a later reuse context and asks whether to persist it, use it locally, verify it, or clarify it with the user. The release contains 140 primary scenarios, 70 evidence-flip contrasts, and a 36-item real-dialogue-derived extension. On 70 adjudicated primary-test items, verification recall exceeds clarification recall in all 12 label conditions across four checkpoints. However, only three comparisons remain significant after an exploratory Fisher test with Holm correction, and none involve Claude. A category oracle attains .800 accuracy, exposing strong scenario-category structure. Original label and tool-selection conditions agree on only 23--57\% of items, although their differing prompts prevent attributing this solely to the interface. The naturalistic extension is descriptive: it contains only two clarification items and evaluates local models only. MCB measures reuse-time decision selection under a conservative policy; storage backends, matched policy/tool interventions, and downstream utility remain unevaluated.
\end{abstract}

\section{Introduction}

LLM agents retain interaction history to personalize later behavior and support long-horizon tasks \citep{packer2023memgpt,chhikara2025mem0}. Memory formation nevertheless requires a decision about scope and authority. A temporary request should not become a standing preference; an old service status may require a current check; and an ambiguous correction may require a question before it changes future behavior.

We study the \emph{memory--clarification boundary}. Given an interaction-derived candidate and a later reuse context, an agent selects one of four actions: \emph{persist}, \emph{ephemeral} use, \emph{verify} against an external source, or \emph{clarify} with the user. Verification and clarification resolve different uncertainties. An external source establishes changing world facts, while the user establishes intent, scope, and ambiguous reference.

The evaluated decision occurs \emph{at reuse time}: the reuse context is already visible to the model. We do not test an acquisition-time filter that predicts which information will be useful before future contexts are known. A reuse-time decision could authorize a subsequent write or defer it, but this benchmark records the decision only. Applications to vector, profile, episodic, or graph memories are architectural proposals rather than demonstrated backend-independent results.

The Memory--Clarification Boundary benchmark (\mcb) contains 140 primary scenarios split into 70 development and 70 held-out items, a separate 70-item contrast extension, and 36 independently adjudicated items derived from public dialogues. We evaluate two Claude checkpoints and two local models, Qwen3.5-9B and Gemma4-E4B, under bare, original policy, and few-shot prompts. \act{} asks for a structured tool call instead of a label. The original policy omits several annotation rules, and the original tool condition has no policy prompt; both facts constrain interpretation.

Our contributions are:
\begin{enumerate}
\item an auditable four-action benchmark with frozen data, annotations, prompts, and item-level outputs supplied as a downloadable supplementary artifact;
\item class-specific and baseline analyses that distinguish observed recall differences from evidence of a general model property; and
\item a comparison of original label and tool-selection settings, identifying sensitivity that motivates a future matched-prompt evaluation.
\end{enumerate}

\section{Reuse-Time Memory Decisions}

Let a candidate update $u$ be extracted from an interaction history $h$ and reconsidered in reuse context $c$. A commitment policy selects
\begin{equation}
g(h,u,c) \in \{P,E,V,C\},
\end{equation}
where $P$ persists the update, $E$ limits it to the current context, $V$ rechecks an external source, and $C$ queries the user. The distinction is based on authority and scope rather than topic words. A restaurant's hours and a user's preferred meeting time may both mention time; the first is world-authoritative and freshness-sensitive, whereas the second may be a durable preference or an ambiguous one-off request.

Figure~\ref{fig:gate} illustrates the evaluated choice and possible downstream handling. The candidate is supplied rather than extracted by the evaluated model; it need not already exist in durable storage. A system could insert or update it after choosing $P$, or defer writing after $V$ or $C$. We do not execute those transitions or assess an existing memory store.

\begin{figure*}[t]
\centering
\small
\setlength{\fboxsep}{5pt}
\begin{tabular}{c@{\hspace{1.1em}$\rightarrow$\hspace{1.1em}}c@{\hspace{1.1em}$\rightarrow$\hspace{1.1em}}c}
\fbox{History and reuse context} & \fbox{Candidate update} & \fbox{Reuse-time decision $g(h,u,c)$} \\
\end{tabular}
\par\vspace{0.55em}
$\downarrow$ route by authority, scope, and freshness
\par\vspace{0.45em}
\begin{tabular}{@{}p{3.55cm}p{3.55cm}p{3.55cm}p{3.55cm}@{}}
\centering\fbox{\parbox{3.15cm}{\centering\textbf{Persist}\\commit to vector, profile, episodic, or graph memory}} &
\centering\fbox{\parbox{3.15cm}{\centering\textbf{Ephemeral}\\use in working context; no durable update}} &
\centering\fbox{\parbox{3.15cm}{\centering\textbf{Verify}\\stage as untrusted; check a world source}} &
\centering\arraybackslash\fbox{\parbox{3.15cm}{\centering\textbf{Clarify}\\stage as unresolved; query the user}} \\
\end{tabular}
\caption{The evaluated reuse-time choice and proposed downstream routes. MCB records action selection; storage writes, source checks, user answers, and backend-specific transitions are not executed.}
\label{fig:gate}
\end{figure*}

The gate is conservative by design. When persistence and a weaker action remain tied, the stated rules prefer the weaker commitment. This is an assumed cost preference: a wrong durable update can affect later behavior, while an unnecessary question is visible. We do not measure these costs or establish that the preference is optimal.

\section{Related Work}

\paragraph{Recall and memory lifecycle.}
LongMemEval, LoCoMo, MemoryBank, and MemBench evaluate long-term conversational memory, knowledge updates, retrieval, and abstention \citep{wu2025longmemeval,maharana2024locomo,zhong2024memorybank,tan2025membench}. Newer benchmarks add real-dialogue memory lifecycles and penalties for obsolete-memory reuse \citep{xiao2026alpsbench,uddin2026memora}. These tasks largely ask whether stored information can be recovered or correctly reused. \mcb{} supplies a candidate update and reuse context and scores the selected response to its scope and authority, without executing retrieval or storage.

\paragraph{Storage, editing, and action.}
PerMemBench learns a binary session-level storage gate \citep{in2026permem}; Memory-R1 learns \textsc{add}/\textsc{update}/\textsc{delete}/\textsc{noop} operations \citep{yan2025memoryr1}; and Mem2ActBench tests whether retrieved memory grounds later tool parameters \citep{shen2026mem2act}. Mem2ActBench evaluates how retrieved memory supports action; \mcb{} scores a supplied candidate at reuse time, including whether it should authorize a durable update. \mcb{} also separates two types of unresolved uncertainty---checking the world and asking the user---that a binary gate conflates.

\paragraph{Structured and graph memory.}
A-MEM dynamically links and evolves interconnected memory records \citep{xu2025amem}, while Zep represents evolving agent memory in a temporal knowledge graph \citep{rasmussen2025zep}. Such systems improve organization, retrieval, and temporal reasoning after information reaches the memory layer. Our focus is complementary: we score a candidate-handling decision, without implementing or comparing either memory architecture. Graph insertion and propagation risks are motivations for future testing.

\paragraph{Clarification and interactive agents.}
CLAMBER documents broad failures to identify and clarify ambiguous needs; related work uses uncertainty or expected value to decide when to ask \citep{zhang2024clamber,testoni2024asking,zhang2025clarify}. Interactive benchmarks such as $\tau$-bench and $\tau^2$-bench show that static answers can diverge from tool-mediated behavior \citep{yao2024taubench,barres2025tau2}. \act{} applies this observation at the memory boundary by replacing labels with structured memory operations.

\paragraph{Grounded operational agents.}

\begin{table*}[t]
\centering
\scriptsize
\setlength{\tabcolsep}{3.2pt}
\caption{Positioning by explicit scored target. A dash means the capability is not a central scored target, not that the dataset can never contain such an event. ``Structured action'' includes tool selection or parameterized memory operations.}
\label{tab:related}
\begin{tabular}{@{}p{5.0cm}ccccc@{}}
\toprule
Benchmark & Recall/reuse & Commitment choice & Ask user & Check world & Structured action \\
\midrule
LongMemEval / LoCoMo \citep{wu2025longmemeval,maharana2024locomo} & $\checkmark$ & -- & -- & -- & -- \\
MemBench \citep{tan2025membench} & $\checkmark$ & -- & -- & -- & -- \\
PerMemBench \citep{in2026permem} & $\checkmark$ & $\checkmark$ & -- & -- & -- \\
CLAMBER \citep{zhang2024clamber} & -- & -- & $\checkmark$ & -- & -- \\
Mem2ActBench \citep{shen2026mem2act} & $\checkmark$ & -- & -- & -- & $\checkmark$ \\
\mcb{} (ours) & -- & $\checkmark$ & $\checkmark$ & $\checkmark$ & $\checkmark$ \\
\bottomrule
\end{tabular}
\end{table*}

\section{Benchmark and Method}

\subsection{Task and Labels}

Each item contains an acquisition context, a candidate update, and a later reuse context. The labels operationalize the stated conservative policy rather than universal ground truth. Under that policy:
\begin{itemize}
    \item \textbf{persist}: explicitly durable preferences, authorized policies, or stable user/organization facts;
    \item \textbf{ephemeral}: information scoped to one artifact, session, or time window;
    \item \textbf{verify}: changing world state or a single noisy signal that must be rechecked; and
    \item \textbf{clarify}: unresolved referent, durability, scope, conflict, or second-hand preference for which the user is authoritative.
\end{itemize}

The seven scenario categories are stable preference, episodic preference, freshness-sensitive fact, one-off correction, policy constraint, ambiguous update, and noisy failure signal. Category identity remains highly predictive of the adjudicated action, as the category oracle in Table~\ref{tab:baselines} demonstrates. Eight lexical traps place cues such as ``always'' or ``today'' in contexts where the usual heuristic action is wrong. Models never receive category names or rationales. Each category contributes 10 primary-test items (14.3\% of that split); Table~\ref{tab:distribution} gives action counts, and Appendix~\ref{app:classdetail} gives their cross-tabulation.

\begin{table}[t]
\centering
\small
\caption{Illustrative \mcb{} decisions.}
\label{tab:examples}
\begin{tabular}{@{}p{2.7cm}p{1.5cm}p{2.2cm}@{}}
\toprule
Candidate and reuse context & Gold & Boundary cue \\
\midrule
``I always prefer dark mode''; configure a new workspace & Persist & Explicitly durable preference \\
``Use APA for this report''; prepare an unrelated report & Ephemeral & One-artifact scope \\
Restaurant is open tonight; rely on it next month & Verify & World state can change \\
``Make it like last time''; several artifacts fit & Clarify & User must resolve referent \\
\bottomrule
\end{tabular}
\end{table}

\subsection{Splits and Independent Annotation}

The 140 primary items are deterministically split 70/70 by sorted identifier within category. The development split supplies four demonstrations, one per action. A separate 70-item contrast extension contains 35 evidence-flip pairs and is used only as a controlled sensitivity analysis.

Two non-authors independently labeled the 70 held-out primary and 70 contrast items while blinded to author labels, categories, rationales, model outputs, and each other. Agreement across all 140 items was 0.971 ($\kappa=0.962$); agreement was 0.943 ($\kappa=0.923$) on the primary test and 1.000 on the contrast set. A blind third non-author labeled only four disagreements. Majority adjudication changed eight primary labels and no contrast labels. The audited held-out distribution is 18 persist, 22 ephemeral, 18 verify, and 12 clarify. These correspond to 25.7\%, 31.4\%, 25.7\%, and 17.1\%, respectively.

\begin{table}[t]
\centering\small
\setlength{\tabcolsep}{4pt}
\begin{tabular}{lrrrr}
\toprule
Split & P & E & V & C \\
\midrule
Development & 20 & 18 & 15 & 17 \\
Primary test & 18 & 22 & 18 & 12 \\
Contrast & 20 & 20 & 15 & 15 \\
Naturalistic & 7 & 21 & 6 & 2 \\
\bottomrule
\end{tabular}
\caption{Gold-action counts: persist (P), ephemeral (E), verify (V), clarify (C). Development labels are author-written; the other splits use non-author adjudication. Primary-test scenario categories each contain 10 items.}
\label{tab:distribution}
\end{table}

\subsection{Real-Dialogue-Derived Extension}

We sample 10,000 rows from a frozen revision of the public non-toxic WildChat corpus \citep{zhao2024wildchat}, without using model outputs. Forty English user turns pass automated privacy/safety screens and a pre-annotation quality review. The user turn is verbatim apart from whitespace; a coordinator paraphrases one candidate update before annotation; and every item receives the same counterfactual reuse context 30 days later. Thus, this is a source-grounded naturalistic check, not a longitudinal memory log.

Two new non-authors independently judge usability and action. Usability agreement is .925; among 35 items both mark usable, action agreement is .743 ($\kappa=.603$). A blind third non-author labels 12 disagreements. Majority vote excludes four unsupported candidates and retains 36: 21 ephemeral, 7 persist, 6 verify, and 2 clarify. Model inference begins only after this freeze.

\subsection{Label and Tool-Selection Evaluation}

Label prompts request JSON with one action. \emph{Bare} provides definitions only; \emph{policy} adds the original five rules; \emph{few-shot} adds four development examples. The original policy states ambiguous referent/scope/durability rules but omits explicit instructions for conflicts, second-hand preferences, and consequential reuse after a preference's scope expires. Its other rules also compress the annotation guide. These are evaluations of the archived prompts, not fully guide-aligned interventions.

\act{} requests one JSON call: \texttt{memory\_write(content)}, \texttt{use\_now(note)}, \texttt{check\_source(query)}, or \texttt{ask\_user(question)}. It describes the tools and supplies no five-rule policy. Tool names map to the four actions; deterministic checks assess minimum argument format and relevance. The harness records selection without executing storage, source, or user effects. Its role framing and wording also differ from label mode, so this comparison does not isolate output format as a causal factor.

Appendix~\ref{app:prompts} documents the original prompts' coverage and a complete guide-aligned policy for label and tool outputs. The aligned policy and an act-plus-policy condition have not been evaluated; no reported score is attributed to them.

\subsection{Metrics and Statistics}

Accuracy measures the proportion of correct decisions. Precision measures how often a selected action is correct; recall measures how often the model selects it when required. F1 combines precision and recall, and macro-F1 weights all four actions equally. Precision is set to zero when an action is never predicted. Over-memory (OM) is erroneous persistence divided by all items. Under-memory (UM) is the proportion of gold-persist items missed; OM and UM therefore have different denominators, 70 and 18 on the primary test. Both appear in Table~\ref{tab:main}.

For primary accuracy and accuracy differences, 2,000 bootstrap resamples quantify uncertainty due to the evaluated items. Paired resampling preserves the same item across compared runs. Exact two-sided McNemar tests count items on which one run alone is correct. Holm correction adjusts for testing multiple comparisons: eight original prompt comparisons and four label--tool comparisons form separate families. We call an adjusted $p<.05$ significant within its family; larger values do not establish absence of an effect. Intervals are unadjusted, so excluding zero need not imply significance after multiple-comparison correction.

For the verify--clarify recall comparison, a two-sided Fisher test compares correct/incorrect counts on 18 verify and 12 clarify items. Holm adjustment covers all 12 label conditions. These tests were added after review and are \emph{exploratory}. Recall-difference intervals resample separately within each gold class. Class difficulty and prompt construction differ, so significance would concern this item set, not a universal model tendency. The small naturalistic extension is reported descriptively, without $p$-values or significance claims.

\section{Experimental Setup}

We evaluate Claude Haiku 4.5 (\texttt{claude-haiku-4-5-20251001}), Claude Sonnet 4.6 (\texttt{claude-sonnet-4-6}), Qwen3.5-9B \citep{qwen35modelcard}, and Gemma4-E4B \citep{gemma4report}. Claude calls are isolated per item; served IDs are recorded.

Both open-weight models run locally through Ollama \citep{ollama} with Q4\_K\_M quantization, temperature 0, seed 13, disabled thinking, and isolated item calls. Qwen has 9.7B parameters; the Gemma checkpoint reports 8.0B total parameters (4.5B effective). Complete model digests, runner versions, prompts, token limits, raw JSON, parsed actions, and token counts accompany every prediction. The Gemma digest is \texttt{c6eb396dbd59...cf982eb}. All 280 Gemma outputs parse successfully.

Reference systems are Always-Persist, a fixed temporal/scope keyword heuristic, Majority Action, and a category-majority oracle. Majority Action uses the evaluated split's most frequent gold action; the oracle uses both test gold labels and category IDs hidden from models. These two are distribution diagnostics, not deployable baselines. The fixed keyword rules receive the same scenario text as models and do not inspect test labels.

\begin{table}[t]
\centering\small
\setlength{\tabcolsep}{3pt}
\begin{tabular}{lccccc}
\toprule
Reference & Acc. & F1 & OM & UM & C/V R \\
\midrule
Always-Persist & .257 & .102 & .743 & .000 & .000/.000 \\
Majority action & .314 & .120 & .000 & 1.000 & .000/.000 \\
Keyword rules & .257 & .228 & .071 & .722 & .833/.056 \\
Category oracle & .800 & .792 & .071 & .167 & .667/.889 \\
\bottomrule
\end{tabular}
\caption{Primary-test references ($n=70$). F1 is macro-F1; C/V R denotes clarify/verify recall. Majority action and category oracle use test labels, and the oracle also sees hidden categories. These two rows are distribution diagnostics, not deployable systems.}
\label{tab:baselines}
\end{table}

The category oracle scores .800 accuracy and .792 macro-F1. Sonnet bare scores .814/.790: these similar results show how much this authored benchmark's category structure explains. Several model conditions exceed .800, but the oracle's access to test labels and hidden metadata prevents interpreting it as a fair trained competitor. Low keyword accuracy (.257) does not establish that the benchmark lacks exploitable structure.

\begin{table*}[t]
\centering\small
\setlength{\tabcolsep}{3pt}
\begin{tabular}{llcccccc}
\toprule
Model & Condition & Accuracy [95\% CI] & Macro-F1 & OM$\downarrow$ & UM$\downarrow$ & Clarify R & Verify R \\
\midrule
Claude Haiku 4.5 & bare & .629 [.514,.743] & .615 & .029 & .167 & .500 & .944 \\
 & policy & .857 [.771,.929] & .846 & .014 & .111 & .750 & .944 \\
 & few-shot & .757 [.643,.857] & .732 & .057 & .056 & .500 & .944 \\
 & act & .514 [.400,.629] & .456 & .143 & .167 & .583 & .778 \\
\midrule
Claude Sonnet 4.6 & bare & .814 [.714,.900] & .790 & .057 & .056 & .500 & .889 \\
 & policy & .843 [.757,.929] & .831 & .014 & .167 & .667 & 1.000 \\
 & few-shot & .814 [.729,.900] & .796 & .043 & .111 & .583 & .944 \\
 & act & .529 [.414,.643] & .531 & .100 & .278 & .500 & .556 \\
\midrule
Qwen3.5-9B & bare & .557 [.443,.671] & .450 & .243 & .000 & .000 & .667 \\
 & policy & .629 [.514,.743] & .542 & .100 & .056 & .083 & .944 \\
 & few-shot & .771 [.671,.871] & .726 & .129 & .000 & .333 & .833 \\
 & act & .343 [.243,.457] & .236 & .057 & .833 & .083 & .056 \\
\midrule
Gemma4-E4B & bare & .629 [.514,.743] & .608 & .171 & .111 & .417 & .722 \\
 & policy & .729 [.629,.829] & .675 & .071 & .167 & .250 & .889 \\
 & few-shot & .714 [.600,.814] & .677 & .143 & .056 & .333 & .833 \\
 & act & .157 [.086,.243] & .116 & .486 & .611 & .000 & .056 \\
\bottomrule
\end{tabular}
\caption{Original held-out conditions on adjudicated gold ($n=70$). OM divides erroneous writes by all 70 items; UM divides missed persist decisions by 18 gold-persist items. Recall (R) uses 12 clarify and 18 verify items. Invalid rate is zero. Act requests one tool and argument from tool descriptions only, without the original five policy rules.}
\label{tab:main}
\end{table*}

\section{Results}

\subsection{Recall Differences and Their Uncertainty}

Verification recall exceeds clarification recall in all 12 original label conditions (Table~\ref{tab:classgap}). After Fisher tests with Holm correction, only Qwen bare ($p_{\rm H}=.004$), Qwen policy ($p_{\rm H}<.001$), and Gemma policy ($p_{\rm H}=.012$) remain significant. No Claude comparison passes this correction. The pattern is descriptive across conditions; it does not establish a general cross-model tendency to prefer verification.

Precision also changes the interpretation. Haiku bare has verify precision .472 versus clarify precision .667; Qwen policy has verify precision .486 versus clarify precision 1.000, but selects clarify correctly on only 1/12 items. High verify recall can coexist with over-selection of that action, and high clarify precision with few opportunities taken. Both classes' precision, recall, and F1 are reported rather than treating recall alone as capability.

Errors remain model-specific: bare Qwen maps seven clarify items to persist and four to verify; Gemma policy maps two to persist, four to verify, and three to ephemeral. Because the original policy omits annotation rules, these counts cannot distinguish prompt coverage failures from intrinsic difficulty. The 12 clarify and 18 verify items also differ in content and difficulty.

\begin{table*}[t]
\centering\small
\setlength{\tabcolsep}{4pt}
\begin{tabular}{llcccc}
\toprule
Model & Prompt & Clarify P/R/F1 & Verify P/R/F1 & $\Delta$ recall [95\% CI] & $p_{\rm H}$ \\
\midrule
Haiku & bare & .667/.500/.571 & .472/.944/.630 & .444 [.139,.750] & .077 \\
Haiku & policy & .692/.750/.720 & .810/.944/.872 & .194 [-0.028,.500] & .274 \\
Haiku & few-shot & .600/.500/.545 & .708/.944/.810 & .444 [.139,.750] & .077 \\
Sonnet & bare & .857/.500/.632 & .762/.889/.821 & .389 [.056,.694] & .102 \\
Sonnet & policy & .800/.667/.727 & .783/1.000/.878 & .333 [.083,.583] & .090 \\
Sonnet & few-shot & .700/.583/.636 & .810/.944/.872 & .361 [.083,.667] & .102 \\
Qwen & bare & .000/.000/.000 & .480/.667/.558 & .667 [.444,.889] & .004 \\
Qwen & policy & 1.000/.083/.154 & .486/.944/.642 & .861 [.639,1.000] & <.001 \\
Qwen & few-shot & .667/.333/.444 & .789/.833/.811 & .500 [.167,.806] & .077 \\
Gemma & bare & .625/.417/.500 & .619/.722/.667 & .306 [-0.056,.639] & .272 \\
Gemma & policy & .750/.250/.375 & .640/.889/.744 & .639 [.333,.917] & .012 \\
Gemma & few-shot & .571/.333/.421 & .833/.833/.833 & .500 [.167,.833] & .077 \\
\bottomrule
\end{tabular}
\caption{Class-specific precision (P), recall (R), and F1 on the primary test. $\Delta$ is verify minus clarify recall; intervals resample within each gold class. Two-sided Fisher tests compare 18 verify and 12 clarify items; Holm adjustment covers all 12 label conditions. Tests are post-review exploratory analyses.}
\label{tab:classgap}
\end{table*}

\subsection{Prompt Effects on Accuracy and Action Rates}

Using the expanded eight-comparison family, Haiku policy is significant relative to bare ($\Delta=+.229$, Holm-adjusted $p_{\rm H}=.003$), and Qwen few-shot remains significant ($\Delta=+.214$, $p_{\rm H}=.002$). Haiku's few-shot gain is not significant after correction ($p_{\rm H}=.070$). Gemma's policy and few-shot gains, $+.100$ and $+.086$, are also not significant ($p_{\rm H}=.898$ for each).

Behavior rates reveal effects hidden by accuracy. Qwen policy reduces OM from .243 to .100 and Gemma policy from .171 to .071, but after the expanded eight-test correction these changes are not significant ($p_{\rm H}=.051$ and $.273$). Both prompts increase verification recall more than clarification recall, but their incomplete guide coverage prevents concluding that a fully specified policy fails to improve clarification.

\subsection{Sensitivity to Tool-Selection Settings}

Table~\ref{tab:act} compares each bare label run to its tool-selection run. Sonnet, Qwen, and Gemma have significant accuracy drops after correction; Haiku does not ($p_{\rm H}=.057$). Raw label--tool agreement is at most .571 and falls to .243 for Gemma. These are not parser artifacts: invalid rate is zero, and every selected Gemma tool includes a payload that passes deterministic well-formedness checks.

Gemma provides the strongest stress test. In label mode it distributes decisions across all four actions and reaches .629 accuracy. In tool mode it calls \texttt{memory\_write} 41 times, \texttt{use\_now} 28 times, \texttt{check\_source} once, and \texttt{ask\_user} zero times. Its accuracy falls by .471 ($p_{\rm H}<.001$), OM rises to .486, and both clarification and verification recall collapse. Valid syntax and minimum argument checks do not establish correct operation choice. Because the label and tool prompts differ, these results demonstrate sensitivity to the two original settings rather than an isolated interface effect.

\begin{table}[t]
\centering
\small
\setlength{\tabcolsep}{5pt}
\caption{Act minus bare label accuracy. $p_{\rm H}$ uses the four-model tool-selection family.}
\label{tab:act}
\begin{tabular}{lccc}
\toprule
Model & $\Delta$ [95\% CI] & $p_{\rm H}$ & Agreement \\
\midrule
Haiku & $-.114$ [$-.214$,$-.014$] & .057 & .571 \\
Sonnet & $-.286$ [$-.414$,$-.143$] & $<$.001 & .571 \\
Qwen & $-.214$ [$-.386$,$-.029$] & .047 & .229 \\
Gemma & $-.471$ [$-.614$,$-.329$] & $<$.001 & .243 \\
\bottomrule
\end{tabular}
\end{table}

\subsection{Controlled Contrast Validation}

The evidence-flip extension provides a controlled check on Qwen prompt effects beyond the primary 70 items. On the combined 140 Qwen items, bare, policy, and few-shot accuracy is .614, .757, and .843. Relative to bare, policy gains .143 and few-shot gains .229 (both $p_{\rm H}<.001$ within the two-test contrast family). Clarification remains the weakest class (.074/.407/.519). Both annotators reproduced all 70 contrast labels, but the examples are rule-authored templates aligned with the policy rules. We therefore retain this set as controlled sensitivity evidence, not naturalistic external validation.

\subsection{Descriptive Naturalistic Check}

Table~\ref{tab:naturalistic} reports Qwen and Gemma; Appendix~\ref{app:naturalistic} gives all 16 local conditions, including exploratory Phi-4-mini and Qwen3-4B runs. Claude was not evaluated on this split. The majority-ephemeral reference reaches .583 accuracy but only .184 macro-F1. No local condition exceeds its accuracy. Qwen's best accuracy is .472; Gemma changes from .167 with bare prompting to .472 with few-shot prompting. These are descriptive differences, not significant effects.

Only two items require clarification. All 16 local conditions miss both. With two such examples, one correct decision would change clarification recall by .50; their failure does not provide a precise population recall estimate or corroborate a cross-model property. Even 0/2 has a 95\% Wilson interval of approximately [0,.658]. The split contains six verify items and 21 ephemeral items, so its class-specific results are also unevenly supported.

Qwen's tool accuracy is .306 versus .472 in bare label mode, with agreement .556. Gemma's tool accuracy is .444 versus .167, with agreement .111: its 36 verify labels become 26 \texttt{memory\_write} calls among the tool outputs, and OM reaches .528. All tool arguments pass the minimum checks. These observations illustrate why accuracy alone can conceal erroneous persistence, but do not establish an effect outside this small local-model sample.

\begin{table}[t]
\centering
\scriptsize
\setlength{\tabcolsep}{3.2pt}
\caption{Descriptive naturalistic results ($n=36$): six verify and two clarify items. F1 is macro-F1. No significance tests are reported; Claude is not evaluated on this split.}
\label{tab:naturalistic}
\begin{tabular}{llccccc}
\toprule
Model & Cond. & Acc. & F1 & OM$\downarrow$ & Ver. & Clar. \\
\midrule
Qwen3.5 & bare & .472 & .381 & .056 & .833 & .000 \\
 & policy & .444 & .352 & .111 & 1.00 & .000 \\
 & fewshot & .472 & .365 & .306 & .500 & .000 \\
 & act & .306 & .177 & .000 & 1.00 & .000 \\
\midrule
Gemma4 & bare & .167 & .071 & .000 & 1.00 & .000 \\
 & policy & .306 & .277 & .028 & 1.00 & .000 \\
 & fewshot & .472 & .411 & .389 & .833 & .000 \\
 & act & .444 & .399 & .528 & .667 & .000 \\
\bottomrule
\end{tabular}
\end{table}

\section{Design Implications and Untested Extensions}

A selected action could be connected to a memory backend, but the present results establish only selection behavior given a reuse context. In a proposed implementation, persist could authorize a durable update, ephemeral could leave the candidate local, verify could stage it pending an external check, and clarify could defer it pending a user answer. These states could be represented in a vector-store metadata layer, profile, episodic record, or temporal graph. We have not compared those implementations or tested the effects on retrieval and downstream utility.

For graph memory, unresolved candidates could carry provenance and restricted retrieval permissions until resolved. Such handling is a design hypothesis, not a demonstrated remedy. A future evaluation should distinguish acquisition-time decisions without known reuse context from the reuse-time decision tested here.

The original label/tool disagreement motivates matching the complete decision rules, examples, and role framing across interfaces, including act-plus-policy. End-to-end testing would additionally execute source checks, ask users, update storage, and measure later task outcomes.

\section{Conclusion}

\mcb{} scores persist, ephemeral use, verification, and user clarification for an interaction-derived candidate at reuse time. On the authored primary test, verify recall exceeds clarify recall in all original label conditions, but only three exploratory comparisons survive correction. The .800 category oracle shows that scenario structure explains much of the attainable accuracy. Original label and tool-selection settings also produce substantially different decisions, with prompt differences limiting causal attribution. The small local-model naturalistic check is descriptive. The benchmark and frozen artifact support auditing these specific findings; evaluating complete policies, matched tool interfaces, larger clarification samples, and downstream memory systems is the next step.

\section{Limitations}

The primary test has 70 authored English items, including 12 clarify and 18 verify cases; each scenario category has 10 items. The category oracle's .800 accuracy reveals strong category/action structure. Class content and difficulty differ, so the recall comparison cannot establish an intrinsic preference for verification. Its Fisher tests are exploratory analyses added after review. The contrast set consists of rule-authored templates, and development demonstrations retain author labels.

The original policy omits conflict, second-hand preference, and expired-scope clarification rules and compresses other guide instructions. Complete guide-aligned label and tool policies are documented but have not been run. The original tool condition differs in wording and role framing as well as output format; no act-plus-policy comparison isolates these factors. We therefore do not claim that full policy prompting fails to fix clarification or that an interface change alone causes the observed differences.

The naturalistic extension has 36 items, only two requiring clarification, and no Claude evaluation. It uses public user turns with coordinator-written candidates and a standardized counterfactual reuse context. It is neither a longitudinal deployment study nor a statistically stable clarification evaluation. Its local-model descriptive results do not validate a property of all four primary-test checkpoints.

We use one serving configuration per checkpoint and one quantized version per local model. The Gemma checkpoint has 8B total and 4.5B effective parameters; comparisons with dense models do not isolate scale. The benchmark records one decision without extracting candidates, executing writes or checks, obtaining user answers, or measuring retrieval and task utility. Reuse context is available at decision time. Acquisition-time gating and claims of representation-independent performance remain untested. The conservative policy also assumes a cost asymmetry that applications may choose differently.

\section{Ethical Considerations}

The primary and contrast benchmarks contain no personal records. The frozen naturalistic extension quotes minimized turns from a public, consent-based research dataset; it excludes records flagged toxic or redacted and omits location, timestamp, IP hash, and request-header fields, but automated de-identification cannot guarantee zero residual privacy risk. Its intended use is to study decisions that may contribute to inappropriate durable memory; we do not measure a deployed reduction in such failures. Nevertheless, a clarification policy can burden users, and a verification policy can expose sensitive queries to external services. Real systems should minimize data disclosure, record provenance and consent, support correction and deletion, and calibrate intervention costs with affected users. Benchmark scores should not be treated as permission to deploy a memory-writing agent without domain-specific privacy, security, and human-factors evaluation.

\section{Declaration on Generative AI}

Generative AI systems were used as experimental subjects and for implementation and language assistance. The authors designed the benchmark, prompts, annotation protocol, analyses, and claims; non-author annotators supplied the reported independent labels. All reported numerical results were generated from frozen item-level outputs by deterministic analysis code and checked against the manuscript. Post-review revision and analysis assistance introduced no new model inference or annotation. The authors remain responsible for the paper's content.

\bibliography{references}

\appendix

\makeatletter
\setlength{\@fptop}{0pt}
\setlength{\@fpsep}{12pt}
\setlength{\@fpbot}{0pt plus 1fil}
\setlength{\@dblfptop}{0pt}
\setlength{\@dblfpsep}{12pt}
\setlength{\@dblfpbot}{0pt plus 1fil}
\makeatother

\section{Reproducibility and Artifact Scope}

The accompanying archive \texttt{NORA2026\_camera\_ready\_supplement.zip} supplies the benchmark and executable reproduction code. It includes all 140 primary items, 70 contrast items, frozen non-author annotation decisions, prompts, checkpoint metadata, item-level predictions, and tests. Naturalistic materials include all 40 screened candidates, the 36 adjudicated items, minimized source provenance, three non-author packets, and all 16 local-model runs. The archive is provided as supplementary material; no public repository URL is claimed here.

Run \texttt{python reproduce\_camera\_ready.py} from the extracted archive using Python 3.9 or newer. Model access is not needed. This validates the frozen data and regenerates the primary baselines, class metrics, recall comparisons, and descriptive naturalistic analysis. The original prompts and outputs remain unchanged; the full-policy specification below is separately versioned and unevaluated. The original manuscript and submission archive are preserved outside the revised release.

\section{Prompt Coverage and Complete Policy}
\label{app:prompts}

The original label policy has five rules: persist explicitly durable preferences, authorized policy-like constraints, or stable personal/organizational facts; keep explicitly scoped instructions ephemeral; verify changing facts or a single noisy failure signal; clarify ambiguous or hedged referent, scope, or durability; and prefer weaker commitment when persistence remains uncertain. It does not explicitly mention conflicts, second-hand preferences, or the annotation guide's consequential expired-scope cases. The original act prompt instead lists the four tools and their arguments, without these policy rules. Appendix material does not retroactively change either evaluated prompt.

For future matched comparisons, the following complete specification covers the guide. It is supplied as executable prompt construction in \texttt{src/aligned\_policy.py}, with identical decision rules across label and tool output formats. No results in this paper use it.
\begin{enumerate}
\item Persist deliberately stated standing preferences, constraints from an authorized source, and stable personal or organizational facts. A durable fact can appear in a local correction; content takes precedence over surface phrasing.
\item Keep clearly artifact-, session-, or time-scoped instructions and local edits ephemeral. A failure clearly caused by temporary context can be retried or left local without a durable belief.
\item Verify changing world state when later reuse exceeds its shelf life, a noisy resource failure when a current probe is the appropriate next step, and second-hand or dated policy claims against an available authoritative source.
\item Clarify unresolved referents, consequential ambiguity about scope or durability, conflicts whose precedence is unclear, second-hand preferences, and expired preferences when consequential reuse makes their continuing validity unclear.
\item For an explicitly ended scope with no consequential uncertainty, choose ephemeral. Distinguish user authority (intent, preferences, referents) from external authority (state, versions, schedules). Prefer weaker commitment when persistence remains uncertain.
\end{enumerate}

\section{Class Distributions and Full Primary Metrics}
\label{app:classdetail}

\begin{table}[t]
\centering\small
\setlength{\tabcolsep}{3pt}
\begin{tabular}{lrrrr}
\toprule
Scenario category & P & E & V & C \\
\midrule
Ambiguous update & 0 & 2 & 0 & 8 \\
Episodic preference & 0 & 9 & 0 & 1 \\
Freshness sensitive fact & 1 & 0 & 9 & 0 \\
Noisy failure signal & 0 & 3 & 7 & 0 \\
One off correction & 2 & 8 & 0 & 0 \\
Policy constraint & 8 & 0 & 2 & 0 \\
Stable preference & 7 & 0 & 0 & 3 \\
\bottomrule
\end{tabular}
\caption{Primary-test category/action cross-tabulation. Each scenario category has 10 items; action labels are not the category labels.}
\label{tab:categorycounts}
\end{table}

\begin{table*}[t]
\centering\small
\setlength{\tabcolsep}{5pt}
\begin{tabular}{llcccc}
\toprule
Model & Condition & Persist P/R/F1 & Ephemeral P/R/F1 & Verify P/R/F1 & Clarify P/R/F1 \\
\midrule
Claude Haiku 4.5 & bare & .882/.833/.857 & .750/.273/.400 & .472/.944/.630 & .667/.500/.571 \\
 & policy & .941/.889/.914 & .947/.818/.878 & .810/.944/.872 & .692/.750/.720 \\
 & few-shot & .810/.944/.872 & .867/.591/.703 & .708/.944/.810 & .600/.500/.545 \\
 & act & .600/.833/.698 & .000/.000/.000 & .700/.778/.737 & .292/.583/.389 \\
\midrule
Claude Sonnet 4.6 & bare & .810/.944/.872 & .857/.818/.837 & .762/.889/.821 & .857/.500/.632 \\
 & policy & .938/.833/.882 & .857/.818/.837 & .783/1.000/.878 & .800/.667/.727 \\
 & few-shot & .842/.889/.865 & .850/.773/.810 & .810/.944/.872 & .700/.583/.636 \\
 & act & .650/.722/.684 & .444/.364/.400 & .833/.556/.667 & .300/.500/.375 \\
\midrule
Qwen3.5-9B & bare & .514/1.000/.679 & .900/.409/.563 & .480/.667/.558 & .000/.000/.000 \\
 & policy & .708/.944/.810 & .900/.409/.563 & .486/.944/.642 & 1.000/.083/.154 \\
 & few-shot & .667/1.000/.800 & .944/.773/.850 & .789/.833/.811 & .667/.333/.444 \\
 & act & .429/.167/.240 & .352/.864/.500 & 1.000/.056/.105 & .125/.083/.100 \\
\midrule
Gemma4-E4B & bare & .571/.889/.696 & .769/.455/.571 & .619/.722/.667 & .625/.417/.500 \\
 & policy & .750/.833/.789 & .810/.773/.791 & .640/.889/.744 & .750/.250/.375 \\
 & few-shot & .630/.944/.756 & .778/.636/.700 & .833/.833/.833 & .571/.333/.421 \\
 & act & .171/.389/.237 & .107/.136/.120 & 1.000/.056/.105 & .000/.000/.000 \\
\bottomrule
\end{tabular}
\caption{Full per-action primary metrics on the same 70 adjudicated items. P/R/F1 denotes precision/recall/F1. No scores use the unevaluated complete policy.}
\label{tab:fullclass}
\end{table*}

Table~\ref{tab:categorycounts} makes category/action structure explicit. Table~\ref{tab:fullclass} reports precision, recall, and F1 for all four actions and all original primary conditions. Wilson intervals for each clarify and verify recall, exact contingency counts, raw Fisher probabilities, Holm-adjusted probabilities, and stratified-bootstrap intervals are included in the supplementary \texttt{results/camera\_ready/analysis.json}.

\section{Full Descriptive Naturalistic Results}
\label{app:naturalistic}

\begin{table*}[t]
\centering\small
\setlength{\tabcolsep}{4pt}
\begin{tabular}{llccccccc}
\toprule
Model & Condition & Accuracy [95\% CI] & Macro-F1 & OM & UM & Ephemeral R & Verify R & Clarify R \\
\midrule
Qwen3.5-9B & bare & .472 [.306,.639] & .381 & .056 & .429 & .381 & .833 & .000 \\
 & policy & .444 [.278,.611] & .352 & .111 & .429 & .286 & 1.000 & .000 \\
 & few-shot & .472 [.306,.639] & .365 & .306 & .000 & .333 & .500 & .000 \\
 & act & .306 [.167,.472] & .177 & .000 & 1.000 & .238 & 1.000 & .000 \\
\midrule
Gemma4-E4B & bare & .167 [.056,.306] & .071 & .000 & 1.000 & .000 & 1.000 & .000 \\
 & policy & .306 [.167,.472] & .277 & .028 & .571 & .095 & 1.000 & .000 \\
 & few-shot & .472 [.306,.639] & .411 & .389 & .143 & .286 & .833 & .000 \\
 & act & .444 [.278,.611] & .399 & .528 & .000 & .238 & .667 & .000 \\
\midrule
Phi-4-mini & bare & .111 [.028,.222] & .071 & .111 & 1.000 & .000 & .667 & .000 \\
 & policy & .194 [.083,.333] & .170 & .278 & .429 & .000 & .500 & .000 \\
 & few-shot & .250 [.111,.389] & .176 & .500 & .000 & .000 & .333 & .000 \\
 & act & .194 [.056,.333] & .096 & .000 & 1.000 & .048 & 1.000 & .000 \\
\midrule
Qwen3-4B & bare & .278 [.139,.417] & .205 & .306 & .286 & .000 & .833 & .000 \\
 & policy & .444 [.278,.611] & .367 & .083 & .286 & .286 & .833 & .000 \\
 & few-shot & .361 [.194,.528] & .275 & .444 & .000 & .000 & 1.000 & .000 \\
 & act & .500 [.333,.667] & .283 & .000 & 1.000 & .571 & 1.000 & .000 \\
\bottomrule
\end{tabular}
\caption{All original local-model naturalistic conditions ($n=36$), reported descriptively. Gold counts are 7 persist, 21 ephemeral, 6 verify, and 2 clarify. No Claude results or $p$-values are reported.}
\label{tab:fullnatural}
\end{table*}

All 16 local conditions use the same 36 items. Qwen and Gemma were selected in the frozen protocol, with Phi-4-mini and Qwen3-4B as exploratory checks. No comparison in this section receives a significance claim; there are only two clarify items. Overall intervals describe variability across this item sample, not uncertainty from repeated model sampling.

\end{document}